\pdfoutput=1

\documentclass[11pt]{article}

\usepackage{emnlp2023}

\usepackage{times}
\usepackage{latexsym}
\usepackage{graphicx}
\usepackage{colortbl}
\usepackage{multirow}
\usepackage{booktabs}

\usepackage{algorithm}
\usepackage{algpseudocode}
\usepackage{float}
\usepackage{hyperref}
\usepackage[nolist]{acronym} 
\usepackage{subcaption}

\usepackage{comment}
\usepackage[T1]{fontenc}

\usepackage[utf8]{inputenc}

\usepackage{microtype}

\usepackage{inconsolata}

\usepackage{graphicx}
\usepackage{latexsym}

\usepackage{xcolor}
\usepackage{array}
\usepackage{tikz}
\usepackage{amsmath}
\usepackage{colortbl}
\usepackage{adjustbox} 
\usepackage{multicol}
\usepackage{longtable}

\title{Classifying multilingual party manifestos:\\Domain transfer across country, time, and genre}

\author{Matthias Aßenmacher$^{1,3}$\textsuperscript{$\spadesuit$}
        \And Nadja Sauter$^1$\textsuperscript{$\diamondsuit$} \\ \\
        $^1$ Department of Statistics, LMU, Munich, Germany \\ 
        $^2$ Munich Center for Machine Learning (MCML), LMU, Munich, Germany\\
        \small \textsuperscript{$\spadesuit$}\texttt{matthias@stat.uni-muenchen.de} \quad \textsuperscript{$\diamondsuit$}\texttt{nadja.sauter@campus.lmu.de} \quad \textsuperscript{$\clubsuit$}\texttt{chris@stat.uni-muenchen.de}
        \And Christian Heumann$^1$\textsuperscript{$\clubsuit$}
        }

\begin{document}

\maketitle

\begin{abstract}
Annotating costs of large corpora are still one of the main bottlenecks in empirical social science research. On the one hand, making use of the capabilities of domain transfer allows re-using annotated data sets and trained models. On the other hand, it is not clear how well domain transfer works and how reliable the results are for transfer across different dimensions. We explore the potential of domain transfer across geographical locations, languages, time, and genre in a large-scale database of political manifestos. First, we show the strong within-domain classification performance of fine-tuned transformer models. Second, we vary the genre of the test set across the aforementioned dimensions to test for the fine-tuned models' robustness and transferability. For switching genres, we use an external corpus of transcribed speeches from New Zealand politicians while for the other three dimensions, custom splits of the Manifesto database are used. While BERT achieves the best scores in the initial experiments across modalities, DistilBERT proves to be competitive at a lower computational expense and is thus used for further experiments across time and country. The results of the additional analysis show that (Distil)BERT can be applied to future data with similar performance. Moreover, we observe (partly) notable differences between the political manifestos of different countries of origin, even if these countries share a language or a cultural background.
\end{abstract}



\section{Introduction}
\label{intro}

Publishing party manifestos in the time frame leading up to an election is a common procedure in most parliamentary democracies around the globe. Summarizing the parties' political agendas for the upcoming electoral period, the published manifestos are intended to serve as guides for voters to reach their decision \citep{Suiter2011}. Since the content of these manifestos also constitutes the foundation for the process of building government coalitions, analyzing them can be very insightful. \citet{doi:10.1177/1354068895001002001}, for instance, investigate the common assumption that political parties often try to change their images following a poor election result. Other researchers examine if parties learn from foreign successful parties \citep{bohmelt2016party}. \citet{tavits2009left} and \citet{tsebelis1999veto} also investigate their research questions based on political manifestos.

The Manifesto Project\footnote{\url{https://manifesto-project.wzb.eu/}} covers programs of over 1000 political parties from more than 50 countries over a time frame from 1945 until today \cite{MP}. The database provides access to the raw content of all documents as well as additional annotation for further analysis. Human annotators from over 50 different countries contributed by splitting the documents into quasi-sentences and subsequently classifying each of them according to a coding scheme covering 54 thematic categories. On a more course-grained scale, these 54 categories were further summarized into eight topics. Since manual annotation is extremely time and labor-intensive, requiring annotator training reliability, (partial) automation of the process could yield enormous potential for savings. 

Our research explores how methods from the field of Natural Language Processing (NLP), which are more and more frequently used in social science research \citep{wankmuller2021introduction}, can be used to classify the quasi-sentences of the political manifestos into the eight topics of the Manifesto coding scheme. Therefore, different NLP methods, namely TF-IDF + logistic regression (LR) as a comparative baseline (cf. \citet{osnabrugge2023cross}) and different monolingual and multilingual versions of BERT \citep{devlin-etal-2019-bert} are used to process and subsequently classify the sequences. In the following, first, the related work (cf. Sec. \ref{sec:related}) and the data extraction process (cf. Sec. \ref{sec:data}) will be explained in further detail followed by the experimental setup (cf. Sec. \ref{sec:exp_setup}), where we delve deeper into the concept of cross-domain classification and motivate the different cross-domain scenarios. The predictive performances of each evaluated model for each of the different scenarios are compared and discussed in Section \ref{sec:res}. We conclude the experiments by fine-tuning a multilingual model on the whole corpus.\\

\noindent \textbf{Contribution:} Our main contributions can be summarized as follows: We extend the cross-domain setting introduced by \citet{osnabrugge2023cross} along multiple axes. We not only measure transfer across genre (manifestos $\to$ speeches) but also across time (2018 $\to$ 2022) and country (leave-one-country-out, LOCO). Instead of relying on simple machine learning classifiers, we fine-tune pre-trained language models \citep{devlin-etal-2019-bert,sanh2019distilbert} achieving superior performance to simple models. We don't only rely on English texts, but leverage the whole Manifesto database by employing multilingual pre-trained models. This enables us to train one single model which can be used for all languages and countries. The code for our experiments and the trained models are publicly available to nurture further research: \url{https://github.com/slds-lmu/manifesto-domaintransfer} (code) and \url{https://huggingface.co/assenmacher} (models).


\section{Materials and Methods}
\label{sec:material}


\subsection{Related work}
\label{sec:related}

We draw inspiration for our work from the research article "Cross-Domain Topic Classification for Political Texts" \citep{osnabrugge2023cross}. The authors employ supervised machine learning (logistic regression, LR) alongside feature engineering techniques for text (TF-IDF w/ n-grams) for the classification of political manifestos and speeches. The analysis was performed on two (labeled) data sets, where each utterance was assigned one of the eight possible categories "freedom and democracy", "fabric of society", "economy", "political system", "welfare and quality of life", "social groups", "external relations" and "no topic". The source corpus consists of manifestos, collected between 1984 and 2018, which were extracted from the Manifesto Project \citep{manifesto2018} for the following seven English-speaking countries: Australia, Canada, Ireland, New Zealand, South Africa, the UK, and the USA. Each document was split into quasi-sentences ($n_{source} = 115,410$) and then labeled by a trained human annotator from the Manifesto Project. In most cases, one quasi-sentence roughly equals one sentence, however, some long sentences containing several statements were split into multiple quasi-sentences. \citet{osnabrugge2023cross} use this source corpus for training and for measuring the within-domain performance. The target corpus ($n_{target}$ = 4,165), consists of English speeches held by members of the New Zealand Parliament in the time period from 1987 to 2002. The speeches were extracted from the official record of the New Zealand Parliament (Hansard), and manually annotated according to the same schema by \citet{osnabrugge2023cross}, who then use it for measuring the cross-domain classification performance.

After the hyperparameter tuning using grid search, they achieve an accuracy of 0.641 on the held-out set of the source corpus and an accuracy of 0.507 on the speeches, showing that cross-domain classification is a reasonable approach. Additionally, the authors create their own, more fine-grained, coding scheme with 44 topic categories for which they report lower performance values for both the within- (0.538) and the cross-domain (0.410) setting. It is important to note, that our performance scores are not perfectly comparable to \citet{osnabrugge2023cross}, since we download the data ourselves (with slight differences, cf. Sec. \ref{sec:data}) and thus have a different train/validation/test split.


\subsection{Data extraction from Manifesto Project}
\label{sec:data}

For conducting the experiments described in Sec. \ref{sec:exp_setup}, we extract the manifestos ourselves from the Manifesto Project database using its dedicated \texttt{R}-package \textit{manifestoR} \citep{manifesto2020r}. Thus, as opposed to \citet{osnabrugge2023cross}, our corpus also includes additional information on the year and country of origin for each utterance. Our data sets include the 2018-2 version of the corpus \citep{manifesto2018}, similar to \citet{osnabrugge2023cross}, as well as the most recent version \citep[2022-1,][]{manifesto2022}, resulting in $n_{2018,en} = 114,523$ for the seven English-speaking countries mentioned in Sec. \ref{sec:related} and $n_{2018,all} = 996,008$ in total. For the 2022 corpus, there are in total $158,601$ English observations and $1,504,721$ for all languages, respectively. Among those, $n_{2022,en} = 27,764$ observations from the period between 2019 and 2022 constitute our test set for the experiments across time for the English language. We observe a difference of 887 samples between the data from \citet{osnabrugge2023cross} ($n_{source} = 115,410$) and our data set ($n_{2018,en} = 114,523$), which is probably due to potential changes in the 2018 version the database.

Figure \ref{fig:cat_corpora} (Appendix \ref{a:label_figs}) visualizes the different label distributions for (a) the source corpus of \citet{osnabrugge2023cross}, (b) our extraction of the 2018-2 corpus, (c) our extraction of the 2022-1 corpus, and (d) the target corpus of the New Zealand speeches \citep{osnabrugge2023cross}. While the former three roughly follow the same distribution, with about 57\% of the observations assigned to either "\textit{welfare and quality of life}" or "\textit{economy}", the most common class of the latter is "\textit{political system}" ($\sim$26\%) followed by "\textit{welfare and quality of life}" ($\sim$19\%). Thus, the two main challenges aside from the domain transfer are the overall class imbalance as well as the differences between the source and target domain with respect to the label distribution. Further Figure \ref{fig:cat_country} (Appendix \ref{a:label_figs}) shows the distribution of the target classes separated by the language the manifestos are written in. We display the three most frequent languages, which we use for conducting experiments across country (cf. Sec. \ref{sec:cross}), against the distribution in the entire 2018-2 corpus of all manifestos. Here we observe some minor differences, as "\textit{welfare and quality of life}" and "\textit{political system}" are more frequently addressed in German-speaking countries (compared to the overall corpus), "\textit{welfare and quality of life}" and "\textit{economy}" in French-speaking ones, and "\textit{political system}" and "\textit{economy}" in English-speaking ones. Notably, for all three languages, the topics "\textit{freedom and democracy}" and "\textit{external relations}" are addressed less often than in the whole 2018-2 corpus.


\section{Experimental Setup}
\label{sec:exp_setup}

In this section, we introduce the concept of domain transfer in general and in particular the cross-domain classification settings for our application. Further, the methodological background for the employed model architectures will be laid out as follows: First, we briefly review common feature engineering techniques for text data and elaborate on the advantages and disadvantages. These techniques include term-frequency inverse-document-frequency (TF-IDF) weighting, as well as dense word or document embeddings. Second, we introduce two state-of-the-art NLP architectures that we employ in our analysis, namely BERT \citep{devlin-etal-2019-bert} and DistilBERT \citep{sanh2019distilbert}, both of which do not require prior feature engineering steps but accommodate the whole pipeline in one single model. Finally, we briefly sketch the individual experiments which were carried out over the course of this study.


\subsection{Cross-Domain Classification} 
\label{sec:cross}

When talking about \textit{classification} in the context of machine learning, researchers commonly implicitly refer to within-domain/within-distribution classification, implying that the trained model is tested on data from the same origin/distribution as the training data (i.e. the \textit{source domain}). Cross-domain classification, on the other hand, explicitly considers a shift in the domain/distribution/source of the data, i.e. the data-generating process is assumed to be different. Frequently examined cases of domain shift in NLP include a change in language (i.e. training the model on text from one language and evaluating it in another, cf. \citet{conneau2018xnli,conneau2019unsupervised}), topic (e.g. training the model on reviews on restaurants and evaluation it on reviews on laptops, cf. \citet{pontiki2014}) or genre (e.g. training on texts and evaluation on transcribed audio data, cf. \citet{osnabrugge2023cross}). In our experiments, we contribute to this body of research by considering the following different cross-domain settings: 


\begin{table*}[ht]
\centering
\caption{Overview of the different cross-domain scenarios investigated in this work, alongside the respectively used corpora, test sets, and examined languages.}
\label{tab:exp}
\begin{minipage}{\textwidth}
\centering
\resizebox{\textwidth}{!}{
\begin{tabular}{lcccccccccc}
\toprule
\multicolumn{1}{c}{} & \multicolumn{2}{c}{Data set characteristica} & \multicolumn{2}{c}{Data set splitting} & \multicolumn{2}{c}{Data set sizes}\\
\cmidrule(l{3pt}r{3pt}){2-3} \cmidrule(l{3pt}r{3pt}){4-5}\cmidrule(l{3pt}r{3pt}){6-7}\cmidrule(l{3pt}r{3pt}){8-9}
Scenario & Corpus & Language(s) & Training set & Test set & Training set & Test set\\
\midrule
within-domain               & 2018-2 & En, De, Fr & random split\footnote{Here: .8/.1/.1, i.e. 80\% of the 2018-2 data.}    & random split\footnote{Here: .8/.1/.1, i.e. 10\% of the 2018-2 data.} & 91,618 / 104,710 / 17,885 & 11,452 / 13,089 / 2,236\\
\midrule\midrule
manifestos $\to$ speeches   & 2018-2 & En         & random split$^{a}$ & speeches  & 91,618 & 4,165 \\
2018 $\to$ 2022             & 2018-2 & En, De, Fr & random split$^{a}$ & future\footnote{\textit{"future"} refers to all the data from the 2022-1 corpus recorded after the 2018-2 cut-off.} & 91,618 / 104,710 / 17,885 & 27,764 / 30,542 / 343 \\
across country              & 2018-2 & En, De, Fr & $n-1$ countries & held-out country & --\footnote{Multiple different scenarios, test set contains one single country in each experiment.} & --$^{d}$ \\
\midrule\midrule
Multilingual                & 2018-2 & 38 languages & random split$^{a}$ & random split$^{b}$ & 796,806 & 99,601 \\
\bottomrule
\end{tabular}
}
\end{minipage}
\end{table*}


\paragraph{Transfer across genre:} We consider party manifestos from all seven (English-speaking) countries as our source corpus $C_{source} = C_{2018,en}$ and evaluate the trained model on a target corpus $C_{target}$ of transcribed parliamentary speeches from New Zealand. This setting is equivalent to the work of \citet{osnabrugge2023cross}, yet we rely on more elaborated model architectures.

\paragraph{Transfer across time:} We use the party manifestos from all countries for all years up until 2018 as source corpus $C_{source}$\footnote{$C_{source}$ is either $C_{2018,en}$, $C_{2018,de}$ or $C_{2018,fr}$}, while the target corpus $C_{target}$ consists of party manifestos from the year 2019 -- 2022. This setting is intended to test the temporal robustness of the fine-tuned models.

\paragraph{Transfer across country:} This setup comprises three distinct experiments for different languages (English, German, French), for each of which we include data from all\footnote{For English we excluded countries with a low $n$, to stay consistent with \citet{osnabrugge2023cross}.} countries, where manifestos in the given language exist in the 2018-2 corpus. The setting for each language consists again of seven (five and four, respectively) different individual experiments, since for each language we include all but one country as source corpus $C_{source}$ and evaluate the model on a target corpus $C_{target}$ including only the manifestos from the single held-out country. Further, we also inspect a true multimodel model trained on data from all available countries.

\paragraph{Metrics and Training} We compare our results, which we measure in terms of Accuracy and Macro-F1 Score, from the cross-domain experiments to the performance we obtain for the within-domain setting. We opt for reporting the macro-averaged version of the F1 Score in order to take into account the class imbalance (cf. Fig. \ref{fig:cat_corpora}). For model training, we conduct a train/validation/test split with proportions .8/.1/.1; all reported performance values are measured on the test set. Note that, depending on the cross-domain setting, also different test sets than the random split are used. Table \ref{tab:exp} summarizes the different investigated scenarios in a comprehensive manner, provides an overview of the respectively used corpora for training and evaluation, and specifies with which procedure the respective test sets were created or selected.


\begin{table*}[ht]
\centering
\caption{Performance values (Accuracy and Macro-F1 Score) of TF-IDF + LR \cite{osnabrugge2023cross} versus English BERT and DistilBERT models (upper part) as well as for German DistilBERT and French FlauBERT models (middle part) and the multilingual DistilBERT model (lover part). Absolute in-/decrease versus the within-domain performance values are appended in parentheses.}
\label{tab:perf_modtime}
\resizebox{\textwidth}{!}{
\begin{tabular}{lcc|cccc}
\toprule
\multicolumn{1}{c}{ } & \multicolumn{2}{c}{within-domain} & \multicolumn{2}{c}{manifestos $\to$ speeches} & \multicolumn{2}{c}{2018 $\to$ 2022} \\
\cmidrule(l{3pt}r{3pt}){2-3} \cmidrule(l{3pt}r{3pt}){4-5}\cmidrule(l{3pt}r{3pt}){6-7} 
& Accuracy & Macro-F1 & Accuracy & Macro-F1 & Accuracy & Macro-F1\\
\midrule
TF-IDF + LR & 0.6413 & 0.5195 & 0.5059 ($\downarrow$ 0.1354) & 0.4474 ($\downarrow$ 0.0586) & -- & --\\
\midrule\midrule
English BERT & 0.6977 & 0.5841 & 0.5613 ($\downarrow$ 0.1364) & 0.5046 ($\downarrow$ 0.0795) & 0.6841 ($\downarrow$ 0.0136) & 0.5707 ($\downarrow$ 0.0134) \\
English DistilBERT & 0.6866 & 0.5694 & 0.5669 ($\downarrow$ 0.1197) & 0.5026 ($\downarrow$ 0.0568) & 0.6784 ($\downarrow$ 0.0082) & 0.5620 ($\downarrow$ 0.0074) \\
\midrule\midrule
German DistilBERT & 0.6583 & 0.5628 & -- & -- & 0.6559 ($\downarrow$ 0.0024) & 0.5485 ($\downarrow$ 0.0143)\\
FlauBERT & 0.6087 & 0.5159 & -- & -- & 0.6093 ($\uparrow$ 0.0006) & 0.4783 ($\downarrow$ 0.0376) \\
\midrule\midrule
Multilingual DistilBERT & 0.6748 & 0.5941 & -- & -- & 0.6311 ($\downarrow$ 0.0437) & 0.5278 ($\downarrow$ 0.0663)\\
\bottomrule
\end{tabular}
}
\end{table*}


\subsection{Model architectures}
\label{sec:models}

Early feature engineering techniques relying on the bag-of-words (BoW) assumption have in recent years been replaced by more elaborated representation learning algorithms. BoW refers to counting the occurrences of words (or n-grams) in a document and representing it as $V$-dimensional vector, where $V$ is the vocabulary size. 
This representation can be enhanced via TF-IDF, as done by \citet{osnabrugge2023cross}, via a re-weighting using corpus-level occurrence statistics.

With the advent of representation learning, it became possible to represent words \citep{mikolov2013efficient,pennington2014glove,bojanowski2016enriching} and documents \citep{le2014distributed} by dense vectors of a comparably low, fixed dimensionality. These representations were used in a similar fashion in conjunction with a classifier as BoW-based representations. BERT \citep{devlin-etal-2019-bert} enabled the coupling of these two steps, i.e. it provided one single end-to-end trainable model for learning (contextual) representations and training the classifier. The commonality of BERT and all subsequent architectures is that they all are relying on the Transformer architecture \citep{vaswani2017attention}. Based on BERT, DistilBERT models can be trained using model distillation \citep{bucilua2006model,hinton2015distilling}, a training process during which the smaller student model (DistilBERT) is trained to mimic the larger teacher model's (BERT) behavior. In the case of DistilBERT, the student model, while having half the size of its teacher model, is able to retain approximately 95\% of the teacher model's performance on the GLUE benchmark \citep{sanh2019distilbert}.

We use \texttt{bert-base-cased} as well as \texttt{distilbert-base-cased} for English. For further experiments, we employ \texttt{distilbert-\allowbreak 
 base-german-cased}, \texttt{flaubert\_small\_cased} (as no French DistilBERT is available) and \texttt{distilbert-base-multilingual-cased}.
 

\subsection{Experiments}
\label{sec:exp}

In the first step, we stick to the setup from \citet{osnabrugge2023cross}, extracting similar data, re-running their experiments, and comparing against their LR+TF-IDF baseline. We further compare the performance of BERT against the cheaper DistilBERT for the English within-domain setting and the English cross-domain settings (manifestos $\to$ speeches, 2018 $\to$ 2022, and across country) to assess the competitiveness of the latter one. For the cross-domain scenarios in the other languages (German, French) we thereafter conduct all experiments with DistilBERT, since it is the cheaper model. The concluding multilingual experiments on the complete corpus are also conducted using a DistilBERT model, fine-tuning the model on the train set of a random split of \textit{the whole} 2018-2 data set.


\begin{table*}[ht]
\centering
\caption{A detailed performance report for per-class within-domain performance, measured in terms of Precision (P), Recall (R), and F1 Score, for the DistilBERT models in English and German, the French FlauBERT as well as for the multilingual DistilBERT. Best scores (per language) in \textbf{bold}, runner-up \underline{underlined}.}
\label{tab:perf_classes}
\resizebox{\textwidth}{!}{
\begin{tabular}{lccc|ccc|ccc|ccc}
\toprule
\multicolumn{1}{c}{ } & \multicolumn{3}{c}{English} & \multicolumn{3}{c}{German} & \multicolumn{3}{c}{French}  & \multicolumn{3}{c}{Multilingual}\\
\cmidrule(l{3pt}r{3pt}){2-4} \cmidrule(l{3pt}r{3pt}){5-7}\cmidrule(l{3pt}r{3pt}){8-10} \cmidrule(l{3pt}r{3pt}){11-13} 
& P & R & F1 & P & R & F1 &  P & R & F1 &  P & R & F1 \\
\midrule
No Topic                    & 0.0000 & 0.0000 & 0.0000 & 0.0000 & 0.0000 & 0.0000 & 0.0000 & 0.0000 & 0.0000 & 0.4142 & 0.1394 & 0.2086 \\
Freedom / Democracy         & 0.6258 & 0.5318 & 0.5750 & 0.6631 & 0.6133 & 0.6372 & 0.6533 & 0.5868 & 0.6183 & 0.6165 & 0.5787 & 0.5970 \\
External Relations          & \textbf{0.7395} & 0.7517 & \underline{0.7456} & \textbf{0.7429} & \underline{0.7067} & \textbf{0.7243} & \textbf{0.6688} & \underline{0.6913} & \textbf{0.6799} & \textbf{0.7357} & 0.7068 & \underline{0.7209} \\
Social Groups               & 0.5794 & 0.5488 & 0.5637 & 0.6040 & 0.5370 & 0.5685 & 0.6034 & 0.4506 & 0.5160 & 0.6242 & 0.5372 & 0.5774 \\
Political System            & 0.5629 & 0.4773 & 0.5166 & 0.6088 & 0.5145 & 0.5577 & 0.4407 & 0.5372 & 0.4842 & 0.6012 & 0.5646 & 0.5823 \\
Fabric of Society           & 0.6463 & 0.6727 & 0.6592 & 0.5909 & 0.6496 & 0.6189 & 0.5485 & 0.4837 & 0.5140 & 0.6212 & 0.6092 & 0.6151 \\
Economy                     & 0.7269 & \underline{0.7570} & 0.7416 & \underline{0.6882} & 0.7009 & 0.6945 & 0.6270 & 0.6449 & 0.6358 & 0.6934 & \underline{0.7449} & 0.7182 \\
Welfare / Quality of Life   & \underline{0.7293} & \textbf{0.7793} & \textbf{0.7534} & 0.6686 & \textbf{0.7379} & \underline{0.7015} & \underline{0.6604} & \textbf{0.6990} & \underline{0.6791} & \underline{0.7151} & \textbf{0.7517} & \textbf{0.7330} \\
\bottomrule
\end{tabular}
}
\end{table*}


\section{Results}
\label{sec:res}

This section will be structured as follows: First, we will show the superior within-domain performance of pre-trained BERT-based models over the simple baseline from \citet{osnabrugge2023cross} and will closely inspect the per-class within-domain performances of the different models. In conjunction with this, we also compare our models to \citet{osnabrugge2023cross} on the manifestos $\to$ speeches scenario, since we adopt it from their work. This scenario we can, however, only inspect for the English language as the corpus of speeches is from New Zealand. Second, we will verify if and how well experiments across genre and time work for the different monolingual models and the multilingual one. Third, we inspect closely how well performance can be transferred across different countries speaking the same language. Subsequently, we delve deeper into a truly multilingual by fine-tuning a pre-trained multilingual model on the entirety of the corpus and examining its performance for the different countries and languages.

\paragraph{Within-domain performance}
The results of our experiments comparing different models for within-domain classification, manifestos $\to$ speeches, and 2018 $\to$ 2022 classification are presented in Table \ref{tab:perf_modtime}. For within-domain classification, the TF-IDF + LR model is clearly outperformed by the deep learning models, where the English models perform better than the German, French, and Multilingual ones. It is notable that in general, the French model exhibits rather low performance values\footnote{Note, that cannot be compared to the English TF-IDF + LR baseline due to different training and test sets.} (within-domain as well as across time) compared to all other models, which may for one reason be caused by the relatively small corpus size for this language compared to all other ones (cf. Tab. \ref{tab:exp}). We also observe the expectedly higher performance of the English BERT model compared to the English DistilBERT, since it generally outperforms DistilBERT in all scenarios except for the accuracy in \textit{manifesto $\to$ speeches} transfer. However, the performance gaps between these two models are rather small, which very well justifies the use of DistilBERT for the remainder of the experiments, trading some performance for saving computational expenses.\footnote{While training BERT for one epoch took roughly 1h 11min, DistilBERT nearly halved this training time per epoch to about 38min. Adding this up over three epochs amounts to time savings of nearly 100min.}

When further considering the predictive performance separately for each of the eight classes (cf. Tab. \ref{tab:perf_classes}), we learn that for none of the languages and for none of the investigated scenarios any of the monolingual DistilBERT models was able to predict a single case of the highly underrepresented "\textit{no topic}" class. The obvious reasons for this are the low number of observations as well as the potential ambiguity, heterogeneity, and fuzziness of the manifestos that could not even by the human annotators be classified into one coherent class but were assigned to this collection basin. This peculiarity of the results should always be taken into account when interpreting them since the macro-averaged F1 Score tends to be a rather conservative performance measure as it weighs the performance of this class similarly to all other classes. This also largely explains the quite notable gap between the Accuracies and Macro-F1 Scores (cf. Tab. \ref{tab:perf_modtime}). 

The largest class (in terms of the number of observations) was easiest to classify for the DistilBERT models across all languages, i.e. for "\textit{welfare and quality of life}" overall the highest values in $P$, $R$, and $F1$ are observed. Interestingly it is not the second largest class ("\textit{economy}") where the models perform next best, but rather one of the smallest classes ("\textit{external relations}"), which is nicely visualized by the highlighting in Table \ref{tab:perf_classes}. Nevertheless, the models are capable of predicting also the "\textit{economy}" class quite well. Further, it is interesting to observe that for the classes exhibiting high F1 Scores, the gap between recall and precision is (a) rather small and (b) sometimes even in favor of the recall, while for the low-performance classes, the recall often appears to be notably worse than the precision. This is especially consistently observable for the class "\textit{social groups}".
 
When compared to the monolingual models, the multilingual one stands out due to two distinct reasons (cf. Tab. \ref{tab:perf_classes}): First, it is the only one of the four models to detect at least \textit{any} true "\textit{no topic}" observations in its test set. Although the performance for this particular class still is not great, it still seems as if learning from more (and more diverse) data seems to help in this respect. Second, and probably also related to the first advantage, the performance seems to be more stable when comparing the scores across the different classes. While for the other English and French, the ranges (excluding "\textit{no topic}") of the F1 Score were 0.2290, and 0.1957 respectively, this metric is with a value of only 0.1556 comparably small, similar to 0.1666 for the German language.

\paragraph{Transfer across genre and time}

Inspecting the two cross-domain settings in Table \ref{tab:perf_modtime} more closely, we see that transfer across the temporal axis works better than across the genre axis. While for the English DistilBERT model the performance on the New Zealand speeches drops by quite a margin ($\downarrow$ 0.1197 / $\downarrow$ 0.0568), it merely changes when evaluated on the data from a different time period ($\downarrow$ 0.0082 / $\downarrow$ 0.0074). Again, comparing BERT to DistilBERT, the latter even seems to be more stable over time since the performance decrease is slightly less pronounced. For the cross-modal transfer scenario, we provide the confusion matrix (cf. Fig. \ref{fig:confmat} in Appendix \ref{a:confmat}) to enable further error analysis. While the two most frequent classes are still very accurately predicted, the model severely struggles when it comes to distinguishing many of the other classes from the "\textit{political system}" category. Even for the two largest classes, a notable amount of the instances were misclassified into this category. Further, the model's error of confusing a certain category with "\textit{political system}" is even worse for the smaller classes, e.g. "\textit{freedom and democracy}", with fewer samples.

While this comparison of the scenarios across genre and across time can not be made for the other languages and the multilingual scenario, we also observe only very minor drops in performance for the latter scenario there. For the two monolingual models, we record decreases for accuracy of 0.24 percentage points for the German model and even no decrease at all for the accuracy of the French DistilBERT model, as well as decreases of 1.43 (German) and 3.76 (French) percentage points for Macro-F1. The multilingual model, however, exhibits somewhat larger drops in performance of 4.37 percentage points for accuracy and 6.63 percentage points for Macro-F1, respectively.


\begin{figure}[ht]
\centering
\includegraphics[width=.48\linewidth]{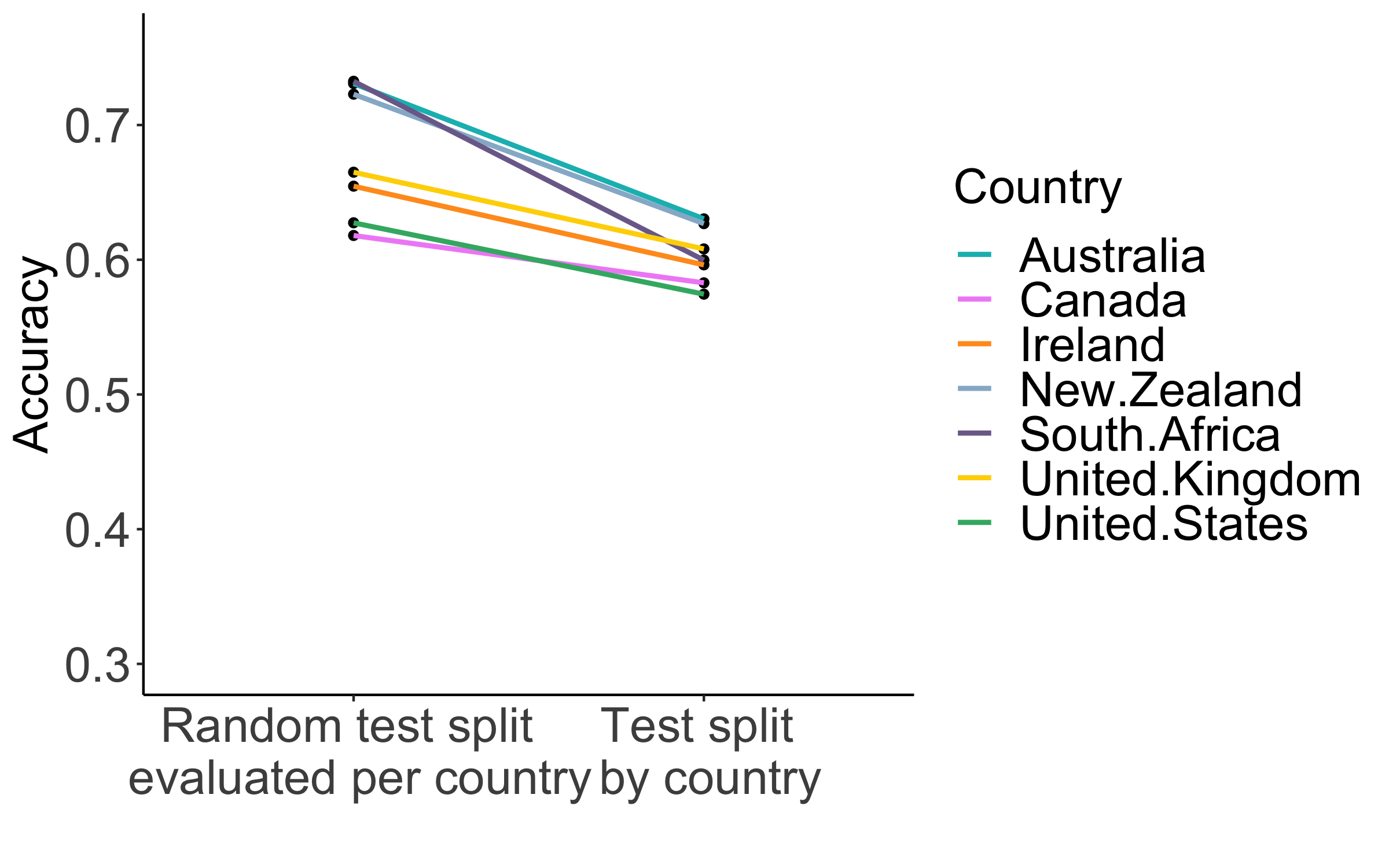}
\includegraphics[width=.48\linewidth]{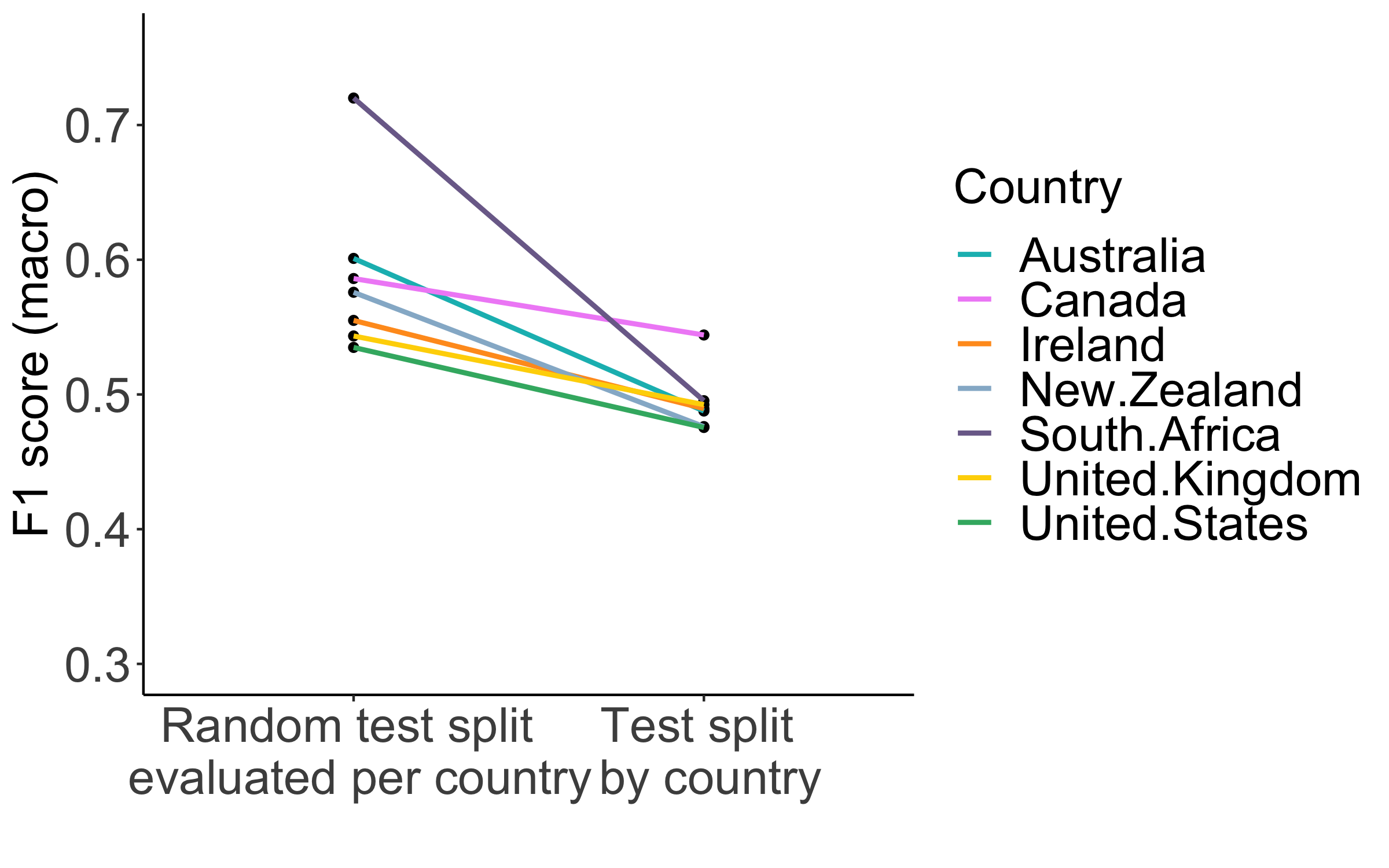}
\includegraphics[width=.48\linewidth]{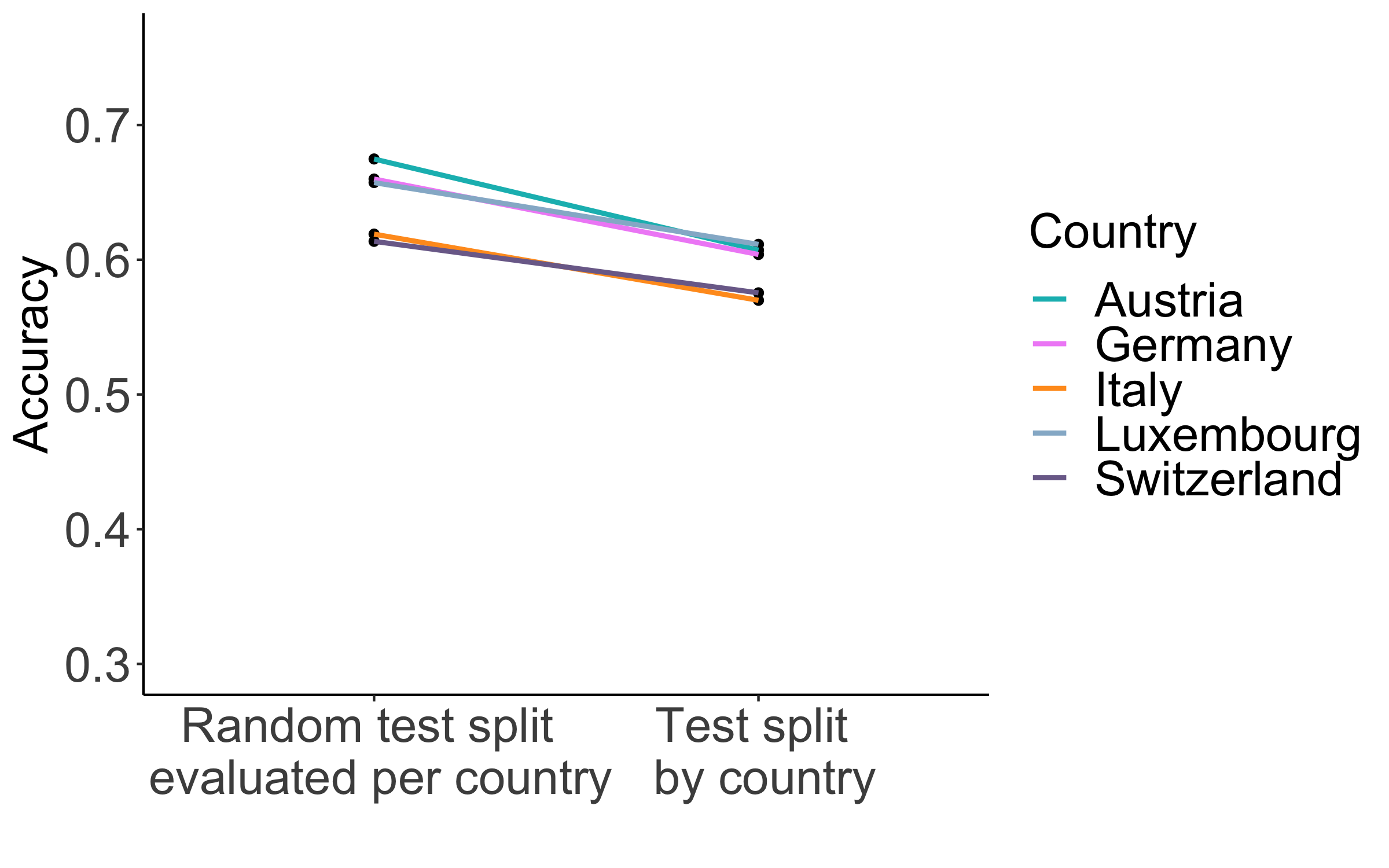}
\includegraphics[width=.48\linewidth]{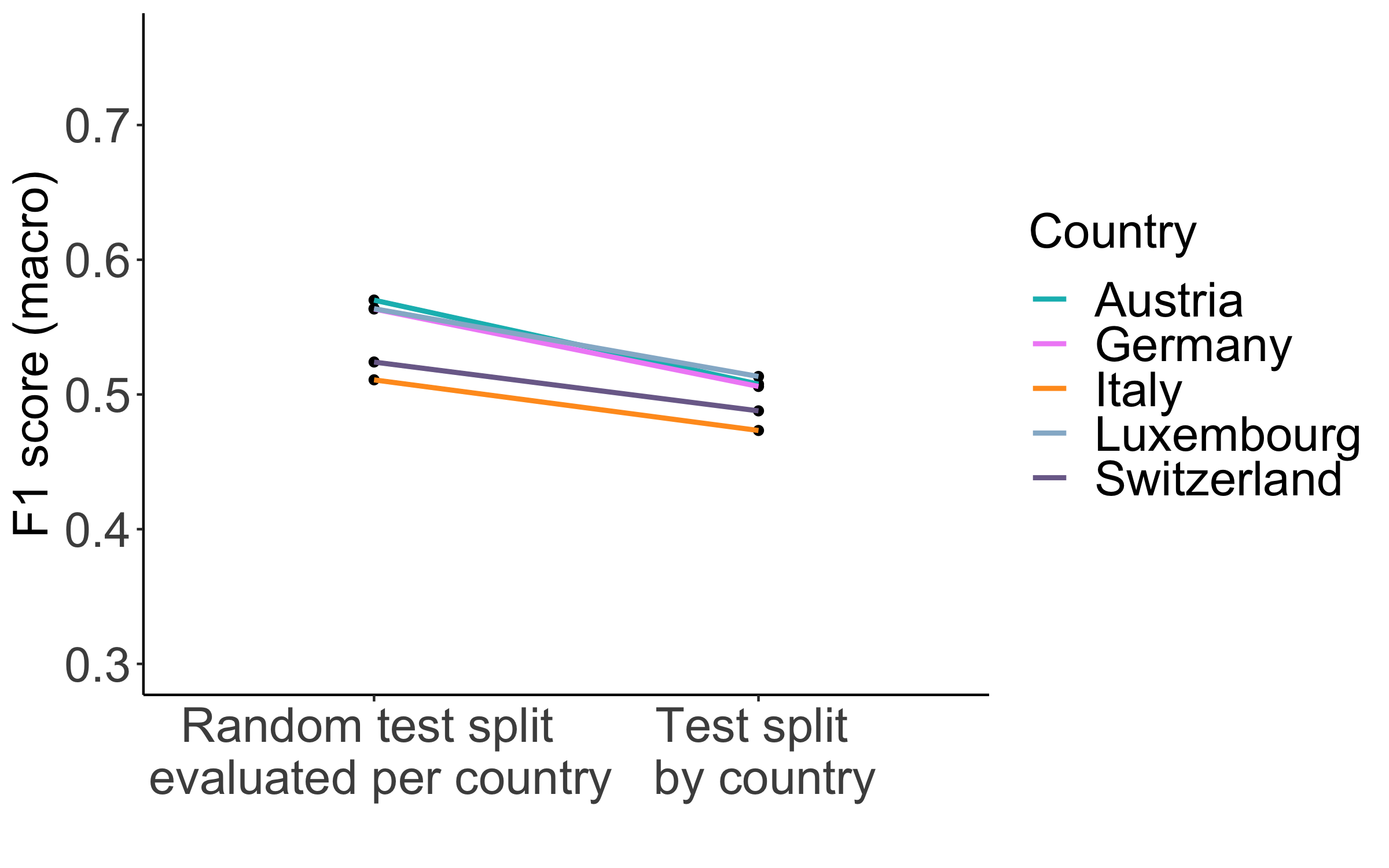}
\includegraphics[width=.48\linewidth]{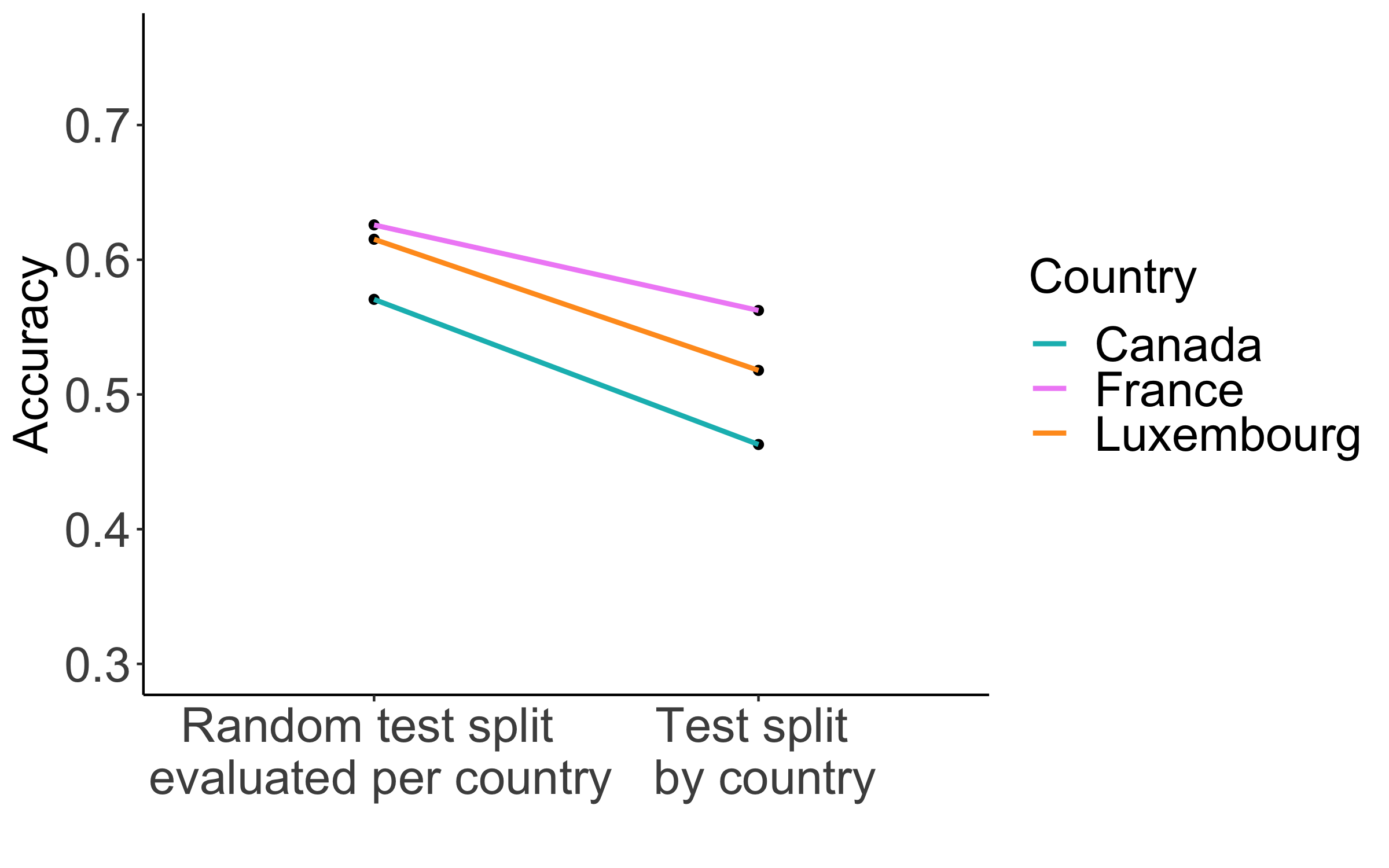}
\includegraphics[width=.48\linewidth]{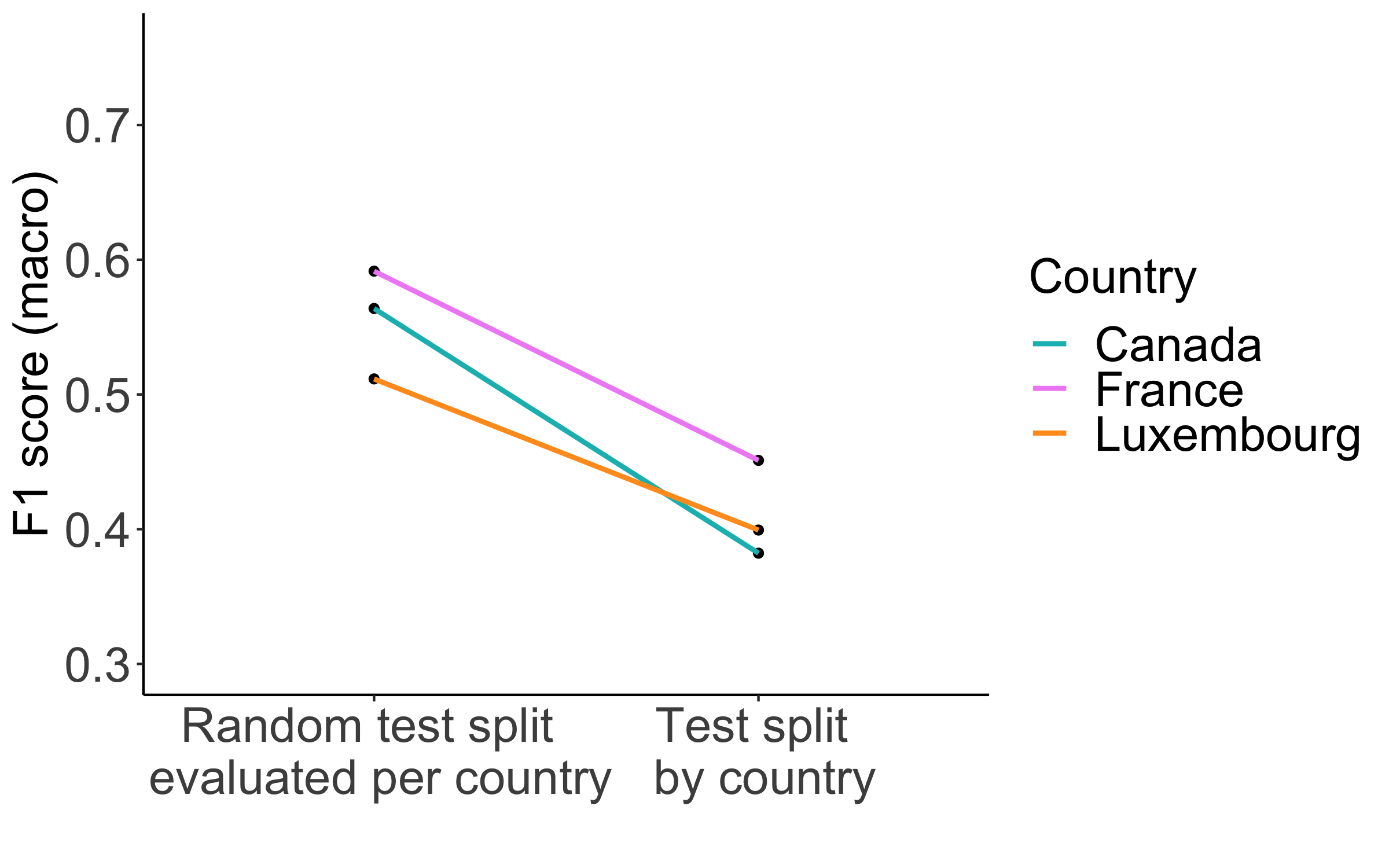}
\caption{Comparison of the performance on data from specific English- (top), German- (middle), and French-speaking (bottom) countries via the Accuracy (left) and Macro-F1. On the left-hand side of each subfigure, performance is measured on the portion of each country in the random test set, while on the right side, the country-specific LOCO performance is displayed. Lines are drawn between the respective points to visualize the connection within one country. Switzerland is excluded, since there is only one sample in the random test split.}
\label{fig:rand_country}
\end{figure}


\begin{table*}[ht]
\centering
\caption{Leave-one-country-out performance (Accuracy and Macro-F1) for English (7 countries), German (5 countries), and French (4 countries). Best scores per language in \textbf{bold}, runner-up \underline{underlined}. We report both $n_{random}$ for the number of observations in the random test split and $n_{country}$ for the number of observations when the respective country is used as held-out set.}
\label{tab:perf_language}
\begin{minipage}{\textwidth}
\resizebox{\textwidth}{!}{
\begin{tabular}{lcc|cc|cc|cc}
\toprule
\multicolumn{1}{c}{ } & \multicolumn{2}{c}{ } & \multicolumn{2}{c}{English-LOCO} & \multicolumn{2}{c}{German-LOCO} & \multicolumn{2}{c}{French-LOCO} \\
\multicolumn{1}{c}{ } & \multicolumn{2}{c}{ } & \multicolumn{2}{c}{(DistilBERT)} & \multicolumn{2}{c}{(DistilBERT)} & \multicolumn{2}{c}{(FlauBERT)} \\
\cmidrule(l{3pt}r{3pt}){2-3} \cmidrule(l{3pt}r{3pt}){4-5} \cmidrule(l{3pt}r{3pt}){6-7}  \cmidrule(l{3pt}r{3pt}){8-9} 
& $n_{random}$ & $n_{country}$ & Accuracy & Macro-F1 & Accuracy & Macro-F1 & Accuracy & Macro-F1\\
\midrule
Australia       & 1,861  & 18,480 & \textbf{0.6304} & 0.4877 & -- & -- & -- & --\\
Canada          & 322    & 3,047  & 0.5829 & \textbf{0.5441} & -- & -- & -- & --\\
Ireland         & 2,548  & 25,357 & 0.5962 & 0.4895 & -- & -- & -- & --\\
New Zealand     & 2,840  & 28,561 & \underline{0.6268} & 0.4761 & -- & -- & -- & --\\
South Africa    & 628    & 6,423  & 0.5997 & \underline{0.4954} & -- & -- & -- & --\\
United Kingdom  & 2,182  & 21,836 & 0.6080 & 0.4924 & -- & -- & -- & --\\
United States   & 1,071  & 10,819 & 0.5744 & 0.4755 & -- & -- & -- & --\\
\midrule
Austria         & 3,361  & 33,818 & -- & -- & \underline{0.6071} & \underline{0.5077} & -- & --\\
Germany         & 6,452  & 63,413 & -- & -- & 0.6039 & 0.5060 & -- & --\\
Italy           & 63     & 651    & -- & -- & 0.5699 & 0.4733 & -- & --\\
Luxembourg      & 1,850  & 19,291 & -- & -- & \textbf{0.6114} & \textbf{0.5134} & -- & --\\
Switzerland     & 1,390  & 13,715 & -- & -- & 0.5754 & 0.4878 & -- & --\\
\midrule
Canada          & 517    & 5,386 & -- & -- & -- & -- & 0.4629 & 0.3822\\
France          & 850    & 8,290 & -- & -- & -- & -- & \underline{0.5624} & \underline{0.4511}\\
Luxembourg      & 868    & 8,662 & -- & -- & -- & -- & 0.5179 & 0.3993\\
Switzerland     & 1      & 19    & -- & -- & -- & -- & \textbf{0.7368} & \textbf{0.7288}\\
\midrule \midrule
\textbf{Average} & & & \textbf{0.6026} & \textbf{0.4944} & \textbf{0.5935} & \textbf{0.4976} & \textbf{0.5700} & \textbf{0.4904} \\
\bottomrule
\end{tabular}
}
\end{minipage}
\end{table*}


\paragraph{Transfer across countries}

The results of our LOCO experiments using the monolingual DistilBERT models for English and German, and a FlauBERT model for French, are presented in Table \ref{tab:perf_language}. We support the results by visualizations (cf. Fig. \ref{fig:rand_country}) of how the performance on manifestos from a certain country changes depending on whether we (a) evaluate on its portion of the random test split or (b) on all manifestos of this country as a hold-out set. The most important takeaway from these illustrations is the fact that completely withholding data from a certain country hurts model performance on data from this specific country, but not in equal parts for the different languages. For German-speaking countries (cf. Fig. \ref{fig:rand_country}, middle) the decrease from left to right is less pronounced than for the other two languages (Fig. \ref{fig:rand_country}, top/bottom). 

The overall takeaway from the previous experiments (better performance for English) is not entirely confirmed by these results, also showing a much more nuanced picture regarding interesting inter-country differences per language. For the LOCO scenario within the English-speaking countries, Australia and New Zealand exhibit the highest values for accuracy, while South Africa and Canada outperform the other with respect to Macro-F1\footnote{Canada has better Macro-F1 Scores than most other countries (except for the top two), but comparably low accuracy.}. The two European countries and the United States overall show the worst performance with respect to both metrics. Further, it is worth noting that there is a rather high variation among these performance values compared to German and French. Excluding the "\textit{no topic}" class, the values for accuracy exhibit a range of 0.0560, while the Macro-F1 Score has a range of 0.0686. On a final note, it is interesting to see that the performance on New Zealand \textit{manifestos} is among the top-ranking countries in accuracy, while the domain transfer across modalities (to New Zealand \textit{parliamentary speeches}) shows a little bit of a performance decrease.

The German LOCO classification experiments using DistilBERT exhibit somewhat different results compared to the English experiments. While the overall averages are comparable, the ranges (0.0415 for accuracy and 0.0344 for Macro-F1) indicate that the values for all countries are relatively similar, with Luxembourg having the highest accuracy of 0.6114 as well as the highest Macro-F1 Score of 0.5134. We speculate that the reason for this observation might lie (a) in the similarity of the political systems\footnote{Despite Luxembourg being a parliamentary monarchy, the country still has a similar landscape of political parties compared to its neighbors, including i.a. social and Christian democrats, liberals, a Green party, as well as different smaller left- and right-wing parties.} of all these countries and (b) in their geographical and cultural closeness. However, being no experts in political science, we would leave the definite interpretation of such matters to those. Regarding the overall performance, the German model performs no worse than the English model(s) which was not necessarily to be expected due to our conclusions drawn from Tables \ref{tab:perf_modtime} and \ref{tab:perf_classes}. 

A rather distinct picture emerges when inspecting the results for the French LOCO classification (still bearing in mind that the performance estimates for Switzerland, with only 19 observations, might make the interpretations rather unreliable). The range for accuracy is 0.2739 and 0.3466 for Macro-F1, which is notably larger than the ranges for both the English-speaking countries and the German-speaking countries. Switzerland exhibits by far the highest values, but it should again be noted that they are based on only 19 observations. The average values are comparable, although a bit lower, to the other two languages, but again strongly influenced by the seemingly strong performance on Swiss manifestos. Regarding the other three countries, France itself stands out from the other two, exhibiting both the highest accuracy as well as the highest Macro-F1 Score among them. 


\section{Discussion and Limitations}
\label{sec:disc}

The advent of large language models (LLMs), in particular ChatGPT \citep{chatgpt,bubeck2023sparks}, resulted in a paradigm change in NLP research. Since then, we can loosely categorize existing and newly introduced classification models into several bins: "pre-train/fine-tune", "prompting", and "chatting"
While "pre-train/fine-tune" has been (and still widely is) the pre-dominant research paradigm in applied NLP research since $\sim2018$, "prompting" has upon the introduction of GPT-3 \citep{brown2020language} become an exciting approach for tackling (a) multi-task learning and (b) low-resource scenarios via few-/zero-shot learning. Further, accessing a model via prompting might be considered more "human-like" / "natural" than training a model on class labels via gradient descent. 

On the other hand, there are still also numerous reasons not to abandon architectures relying on the "pre-train/fine-tune" paradigm \citep{yang2023harnessing}, several of which we consider fulfilled as far as our research question is concerned. First, given the large, annotated training corpus there is no need to rely on few-shot learning but rather to use all of the available data points to achieve maximum model performance. Prompting models would struggle with this amount of data due to context length constraints. Second, given the very custom-defined label set of political topics for this political corpus, for general-purpose prompting models, this label set would always have to be in some way appended to the prompt for the model to be informed about the granularity in the first place. On the one hand, this would probably lead to the model struggling with learning the underlying concepts, on the other hand, it would lead to better adaptive capabilities in case the granularity changes. Third, for domain-specific research questions like this, it might not always be feasible for researchers to access the computational resources for running or prompting such large models, and hence a task-specific, parameter-efficient model that does the trick equally well might be preferable.

We further acknowledge that the performance could potentially still be increased using more elaborate models following the "pre-train/fine-tune" paradigm, e.g. variants of the T5 model family \citep{raffel2020exploring,xue2020mt5}. Using these models, however, come at the cost of a higher computational expense potentially requiring much more VRAM than the average practitioner has access to. The models we employ can, on the other hand, be fine-tuned comfortably using smaller GPUs with around 16GB of VRAM in an acceptable amount of time. Given the ever-increasing model sizes and thus also the computational requirements, this is an important issue to keep an eye on.


\section{Conclusion and Future Work}
\label{sec:concl}

We showed in a series of extensive experiments that domain transfer along three different axes (genre, time, country) in principal works for this sort of political text. We observed the largest performance drops when attempting to generalize across modalities, however, the models tend to generalize very well across time. While the first finding might be foreseeable, the latter result is insofar kind of interesting since after the time point we chose for splitting the data (2018) quite some new topics, e.g. the global covid-19 pandemic or the Ukrainian war, emerged. Regarding the generalization across country, even within languages (and hence to some extent also cultural backgrounds), there seem to be notable differences between the political communication in the different countries as observed by the large performance differences. To conclude, we can state that a true multilingual approach towards classifying political text looks promising, yielding good and stable performance across numerous countries with different languages.

Interesting starting points for future work are obviously to examine the capacities of the emerging ever more powerful LLMs to tackle challenging tasks like this and to make use of the continuously extending data pool from the Manifesto project. Since new countries and time points are added constantly, there is definitely the potential to extend our work in future research.

\section*{Ethical considerations}

To the best of our knowledge, no ethical considerations are implied by our work. The only aspect that is affected in a broader sense is the environmental impact of the computationally expensive experiments. This issue naturally comes with pre-training large language models and is obviously a concern that has to be expressed in every work dealing with this sort of model. But on the other hand, our work rather works against increasing the environmental impact, since we "only" focus on reusing existing pre-trained models and performing the cheap(er) fine-tuning step. Further, we also provide access to our fine-tuned models which can be used by other researchers.

\section*{Acknowledgements} 

This work has been partially funded by the Deutsche Forschungsgemeinschaft (DFG, German Research Foundation) as part of BERD@NFDI - grant number 460037581.



\bibliography{literature}
\bibliographystyle{acl_natbib}


\clearpage

\appendix

\section{Label distributions}
\label{a:label_figs}

\begin{figure}[ht]
\centering
\includegraphics[width=.37\textwidth]{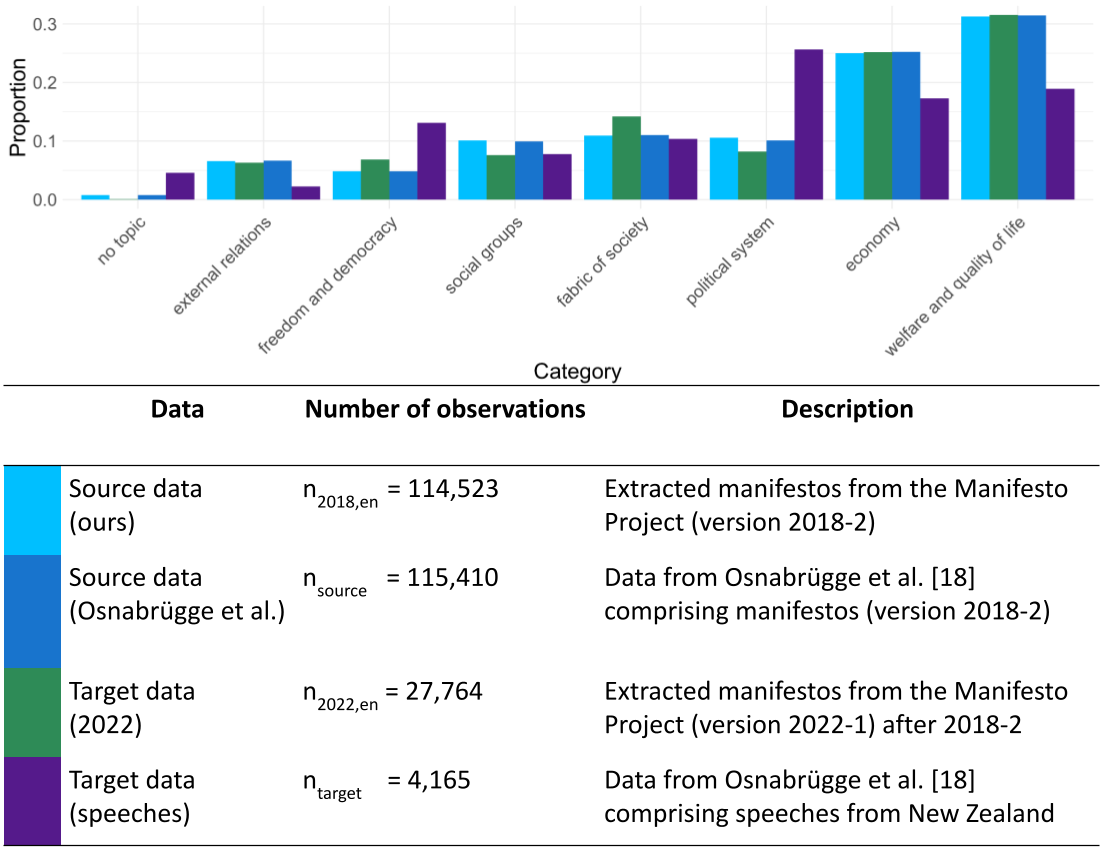}
\caption{Label distributions for the four different corpora alongside sample sizes and short descriptions.}
\label{fig:cat_corpora}
\end{figure}

\begin{figure}[ht]
\centering
\includegraphics[width=.37\textwidth]{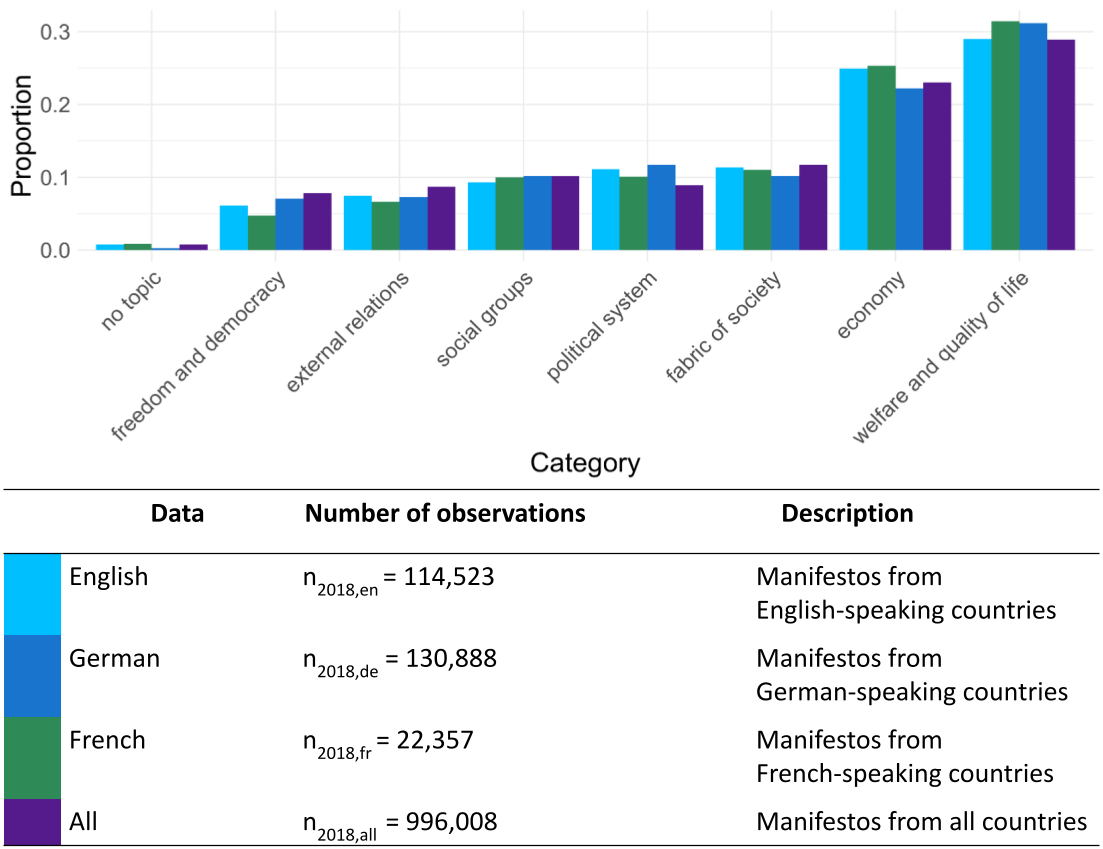}
\caption{Label distributions for the three most frequent languages and overall in the 2018-2 corpus alongside sample sizes and short descriptions.}
\label{fig:cat_country}
\end{figure}


\section{Confusion matrix}
\label{a:confmat}

\begin{figure}[ht]
\centering
\includegraphics[width=.37\textwidth]{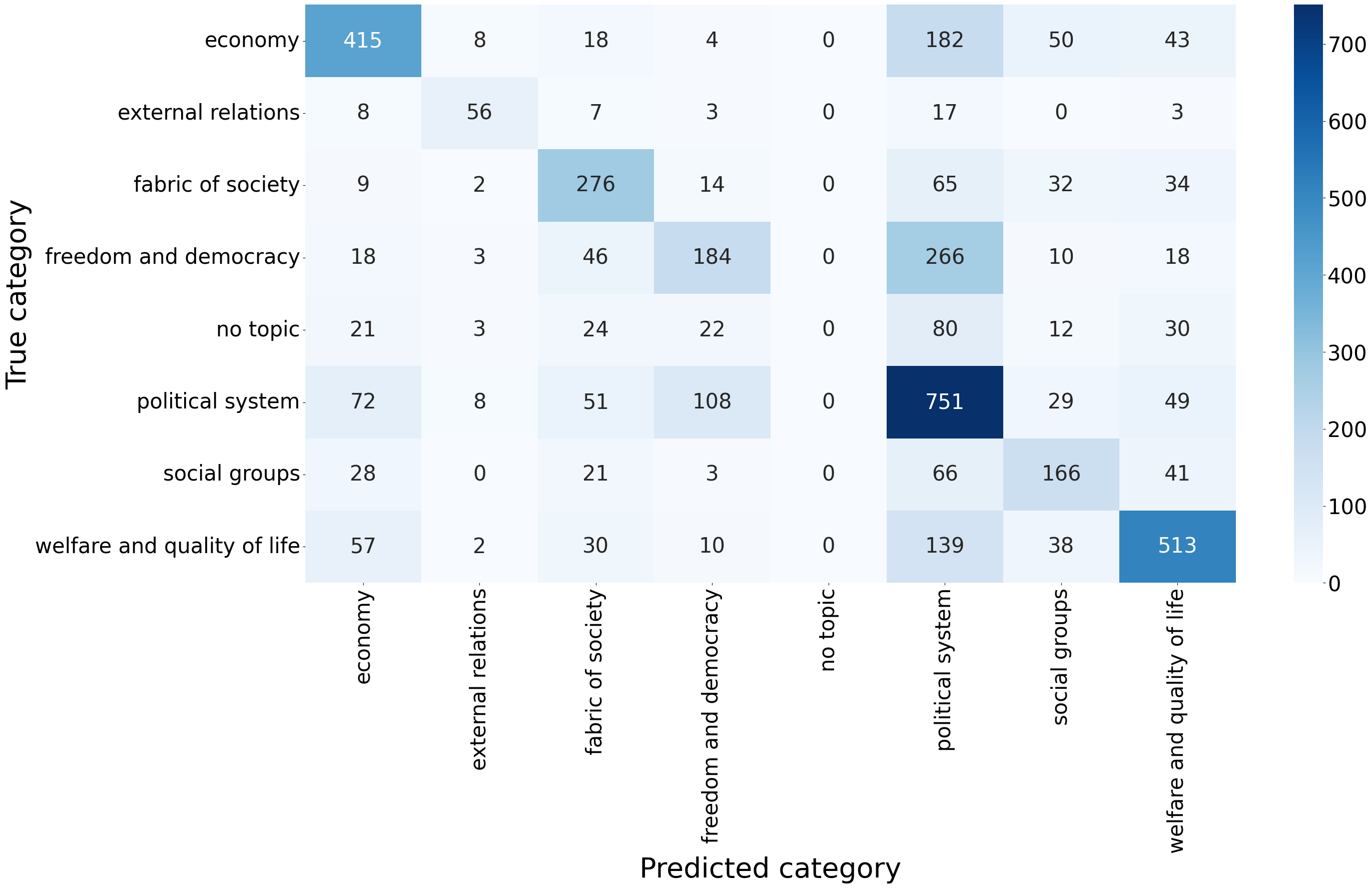}
\caption{Confusion matrix for the performance of the English DistilBERT model on the test set of the New Zealand parliamentary speeches.}
\label{fig:confmat}
\end{figure}

\end{document}